\pdfoutput=1
\documentclass[10pt]{article}

\newcommand{\cv}[1]{}
\newcommand{\av}[1]{#1}

\usepackage{booktabs}
\usepackage{siunitx}
\usepackage{longtable}
\usepackage{tikz}
\usetikzlibrary{arrows.meta}

\av{
\usepackage[T1]{fontenc}
\usepackage{amsmath,amssymb}
\usepackage{graphicx}
\usepackage[hyphens]{url}
\usepackage{natbib}
\bibliographystyle{plainnat}
\usepackage{geometry}
\geometry{letterpaper,width=5.5in,height=8.5in,top=1in,headheight=14pt}
\usepackage[hidelinks]{hyperref}
\urlstyle{rm}
\hypersetup{%
  pdftitle={Neurosymbolic Discovery of Algebraic Graph Constructions},%
  pdfauthor={David Seka, Stefan Szeider},%
  pdfkeywords={neurosymbolic AI, graph theory, computer algebra, large language models, MCP}}
\newcommand{\figureref}[1]{Figure~\ref{#1}}
\newcommand{\tableref}[1]{Table~\ref{#1}}
\newcommand{\sectionref}[1]{Section~\ref{#1}}
\newcommand{\floatconts}[3]{\centering #2\label{#1}#3}
\newcommand{\acks}[1]{\section*{Acknowledgments}#1}
\date{}
}

\newcommand{\Lang}{\mathcal{L}}        %
\newcommand{\Templ}{\mathcal{T}}       %
\newcommand{\Aut}{\operatorname{Aut}}
\newcommand{\eval}{\operatorname{eval}}
\newcommand{\Cay}{\operatorname{Cay}}

\cv{
\title[Neurosymbolic Discovery of Algebraic Graph Constructions]{Neurosymbolic Discovery of Algebraic Graph Constructions}

\author[Seka \& Szeider]{\Name{David Seka} \Email{dseka@ac.tuwien.ac.at}\\
 \Name{Stefan Szeider} \Email{sz@ac.tuwien.ac.at}\\
 \addr TU Wien, Vienna, Austria}
}

\av{
\title{Neurosymbolic Discovery of\texorpdfstring{\\}{ }Algebraic Graph Constructions\texorpdfstring{\thanks{To appear in the Proceedings of the 20th Conference on Neurosymbolic Learning and Reasoning (NeSy 2026), Proceedings of Machine Learning Research (PMLR), vol.~284.}}{}}

\author{David Seka \qquad Stefan Szeider\\[4pt]
  \small Algorithms and Complexity Group\\[-3pt]
  \small TU Wien, Vienna, Austria\\[-3pt]
  \small \texttt{\{dseka,sz\}@ac.tuwien.ac.at}}
}

\begin{document}

\maketitle
\av{\thispagestyle{empty}}

\begin{abstract} %
There are several methods for searching for graphs with prescribed properties, such as SAT solvers and specialized generators.
These methods return the result as raw data: an adjacency matrix or an equivalent string encoding.
The raw data certifies that the graph exists, but it does not reveal any structural properties of the graph: no algebraic construction or decomposition.
We ask whether one can automatically discover a short algebraic description if only this raw data is provided.
We look for a description such as a Cayley graph $\Cay(\Gamma, S)$ or a lexicographic product $C_5[K_3]$.
The description must be precise such that one can create an isomorphic copy of the given graph from the description.

We address this question by means of a neurosymbolic approach.
We propose an agent that runs on a general-purpose large language model with no fine-tuning or per-target training.
The model interleaves reasoning with calls to a computer algebra system.
This way it analyzes the target graph and proposes and tests candidate constructions.
The agent runs the construction and revises it until the output matches the target.
The agent communicates with the computer algebra system through a Model Context Protocol (MCP) server.
We release this server as a general-purpose bridge between agents and the computer algebra system SageMath.
Whether the construction matches the target is checked by a single exact isomorphism test, and therefore rests on the symbolic side and not on the model.
Wrong proposals are rejected, not trusted.
We test the approach on a benchmark of 100 highly symmetric graphs, namely two-orbit graphs on up to 25 vertices, that are likely to allow an algebraic construction; the benchmark was fixed in advance.
Our agent could find verified algebraic constructions for all of them, without falling back to raw encodings.
A strong template-enumeration baseline reaches only about $20\%$.
Also, a catalog lookup could not identify any of these graphs.
However, construction quality declines when symmetry is removed, as expected.
The agent returns the raw encoding only when no symmetry-exploitable structure is available.

As a concrete application of our approach, we identify the smallest known counterexample to the Bernhart--Kainen dispersability conjecture, a $16$-vertex graph that enumeration found as raw data.
For this graph, our agent found an explicit algebraic construction.
\end{abstract}

\section{Introduction}
\label{sec:intro}

A standard technique in combinatorics is to search for a graph with a prescribed property by means of a generator or a SAT solver.
If the graph is found, it is available as raw data.
For instance, exhaustive generators based on canonical augmentation~\citep{McKay1998,McKayPiperno2014} enumerate every graph of a given order and output them as \texttt{graph6} strings.
For eleven vertices, such a catalog already contains more than a billion graphs.
Specialized generators can limit the search to special graph classes like cubic graphs~\citep{BrinkmannEtAl2011}.
On the other hand, SAT-based search can be used to find graphs satisfying properties encoded in propositional logic.
Examples of such approaches have been proposed by \citet{CodishEtAl2013} using symmetry-broken search, by the SAT modulo symmetries tool~\citep{SMS2021}, and by the SAT+CAS tool~\citep{BrightEtAl2019}.
All these approaches have in common that the output is an object of the same kind: an adjacency matrix, a \texttt{graph6} string, or an isomorphism certificate.
This output proves existence, but it does not provide any insight into the structure of the graph.
Two graphs with similar raw data might be very different objects.
One could be a vertex-transitive circulant, the other a structureless random regular graph.
Just looking at the raw data does not distinguish these graphs.
However, an algebraic description such as a Cayley graph $\Cay(\Gamma, S)$ exposes the automorphism group order, regularity, diameter, and structural invariants.
This explains what the graph is.

Just comparing the raw data to a standard catalog or a template-matching approach is insufficient.
On a benchmark of $100$ graphs (\sectionref{sec:experiments}), template enumeration over a hand-built grammar covered only $21$ graphs.
Literature lookup against the House of Graphs database~\citep{BrinkmannEtAl2013} together with recognition checks for circulants, strongly regular graphs, Kneser graphs, Paley graphs, and line graphs could not identify a single one of these hundred graphs.
The remaining $79$ graphs have constructions that are outside the predefined grammar.
As we shall see, these graphs also have algebraic structure, which our approach revealed.

We therefore propose a neurosymbolic approach to the discovery of a construction that goes beyond retrieving one from a catalog or a fixed template.
Our LLM-based agent discovers the construction by analyzing the target graph and testing hypotheses by means of a computer algebra system.
The agent follows the ReAct pattern of interleaved reasoning and action~\citep{ReAct2023}.
The agent can read various properties of the graph under investigation, like degrees and orbit structure, and make hypotheses for a construction.
It can test the hypothesis via the computer algebra system to check against the given target graph.
The agent tries different constructions, revises and corrects errors that surface on execution, until it finds a graph that matches the target.
If the agent cannot find a construction, it can fall back to a construction that replicates the raw data.
For the connection between the agent and the computer algebra system SageMath, we created a dedicated MCP (Model Context Protocol) server.

The correctness of the construction is certified from the symbolic side, by the computer algebra system, which functions as the \emph{trust base}.
A wrong hypothesis is rejected: the model proposes but is never trusted.
We also verify the constructions independently with the computer algebra system after the agent has terminated its work.
Our contribution is therefore the pairing of a neural proposer, which is not based on any fixed template or grammar, with a symbolic verifier that makes the result reliable.
This design does not depend on a particular LLM.
For our experiments, we use a general-purpose frontier model, run at temperature~$0$ (\sectionref{sec:method}).
However, the choice of the model does affect how often and how well constructions are found.
We compare candidate models (\sectionref{sec:backbone}).
We release the MCP server as a standalone contribution that provides any MCP-compatible agent access to the SageMath computer algebra system (\sectionref{sec:method}).
We also release the benchmark with the baselines.

For each admissible order $n \le 25$ (defined in \sectionref{sec:bench-pop}), the benchmark takes the ten connected two-orbit graphs with the largest number of automorphisms, restricted to regular non-joins.
This gives $100$ graphs in total.
These are fixed in advance.
These hundred graphs are highly symmetric on purpose, so that it is reasonable to assume that some algebraic construction exists.
On these hundred graphs, the agent found verified constructions for all of them, not using the fallback to explicit constructions.
The strongest baseline only reached $21$ of these graphs.
We rated the constructions in terms of algebraic canonicity with an LLM judge on a scale of zero to five (\sectionref{sec:human-calib}).
The hundred constructions achieved a mean of $4.4$.
In contrast, we tested the agent on $50$ random regular graphs with $16$ vertices, stratified by the number of automorphisms.
As expected, the construction-quality grade falls monotonically from $5.0$ for the most symmetric graphs to $2.3$ for the least symmetric.
In that group, the trivial raw-encoding fallback appears only for the graphs with low symmetry.
On a set of twenty graphs with planted constructions, the agent recovered nineteen (\sectionref{sec:recall}).
Therefore, overall, the fallback reflects the absence of structure and not that the search failed.

We provide the agent's process for a single example in full in \sectionref{sec:experiments}.
It is a $16$-vertex two-orbit graph, which we identified as the smallest known counterexample to the Bernhart--Kainen dispersability conjecture~\citep{BernhartKainen1979,AlamBekosDujmovicGronemannKaufmannPupyrev2021}.
It improves upon the $20$-vertex Folkman graph, the previously smallest known counterexample.
We found the graph by searching through the catalog of two-orbit graphs we published recently~\citep{seka2026enumeratingtwoorbitgraphs}.
With our agent, we could determine an algebraic construction of this graph.

\section{Task Formalization}
\label{sec:setup}

The task is to turn a graph into a construction.
The input to the task is an undirected simple graph $G$, given as raw data, i.e., an adjacency matrix (in practice, the equivalent \texttt{graph6} string).
The output is a short program $p$ in SageMath (a computer algebra system; henceforth Sage), assembled from standard graph-construction primitives, whose result $\eval(p)$ is isomorphic to $G$; we call such a $p$ a \emph{construction} for $G$ and say it is \emph{accepted}.
Isomorphism is the only correctness requirement, and Sage's method \texttt{is\_isomorphic} decides correctness.

The agent accesses Sage via the MCP protocol and has all the Sage primitives and operations at its disposal. 
This includes a library of algebraic structures, named families of graphs such as cycles $C_n$, paths $P_n$, complete graphs $K_n$, complete bipartite $K_{a,b}$ and multipartite $K_{a_1,\ldots,a_k}$ graphs, and graph operations such as complement, line graph, subdivision, join $\vee$, disjoint union $\sqcup$, and the Cartesian $\square$ product.
This is not an exhaustive list.
In our experiments, the agent mainly used groups to construct the target graphs.
A construction is any well-formed Sage expression that builds a graph.
We write $\Lang$ for this open \emph{construction language}.

A graph will have many possible constructions.
Among those, we want a simple and elegant one, ideally one that exhibits structure, such as a $5$-cycle with every vertex replaced by a triangle.
At the other extreme, a construction can spell out the graph edge by edge.
Although this is a valid construction, we refer to it as a \emph{fallback}: the program \texttt{Graph(g6($G$))} reads the graph from its raw data.
Here we keep the term ``simplest'' informal rather than giving a fixed definition.
We grade accepted constructions for simplicity and elegance in a further process (\sectionref{sec:human-calib}).

As a baseline, we define a subset $\Templ \subseteq \Lang$ of standard templates.
It contains all constructions built from cycles, complete, and complete multipartite graphs by at most three nested operations, plus a fixed list of named schemas: $K_{a,b}$ with a decorated remainder, two cliques with an apex, prisms and cycles attached to a $K_{a,b}$, lexicographic products with edgeless factors, crown graphs, Cartesian products, and complements of these.
Whether $p \in \Templ$ can be decided by inspecting its expression tree.

Sage's \texttt{is\_isomorphic} implements partition-refinement canonical labeling, the exhaustively tested technique behind \texttt{nauty} and \texttt{Traces}~\citep{McKayPiperno2014}; at $n \leq 25$ it runs in well under a second per call, and its correctness on these instances is unconditional.
SageMath provides the \emph{trust base}.
Every $p$ reported as accepted is certified by SageMath.
This is deterministic and does not involve the language model.

\section{Method}
\label{sec:method}

\begin{figure}[t]
\centering
\resizebox{\linewidth}{!}{%
\begin{tikzpicture}[
  font=\small,
  box/.style={draw, rounded corners=3pt, align=center, inner xsep=8pt, inner ysep=6pt, thick},
  databox/.style={box, fill=blue!8, draw=blue!55!black},
  procbox/.style={box, fill=blue!3, draw=blue!55!black},
  trustbox/.style={box, fill=green!10, draw=green!45!black},
  arc/.style={-{Stealth[length=2.6mm]}, line width=0.9pt, black!45},
  barc/.style={{Stealth[length=2.6mm]}-{Stealth[length=2.6mm]}, line width=0.9pt, black!45},
]
\node[databox]  (input)  at (0, 1.0)  {Input:\\ \texttt{graph6} string};
\node[databox]  (output) at (0, -1.0) {Output:\\ algebraic construction};
\node[procbox]  (agent)  at (4.6, 0)  {ReAct Agent\\ (LLM)};
\node[procbox]  (mcp)    at (8.2, 0)  {MCP server\\ (\texttt{mcp-sage})};
\node[trustbox] (sage)   at (11.4, 0) {SageMath\\ (trust base)};

\draw[arc] (input.east)  -- (agent.170);
\draw[arc] (agent.190)   -- (output.east);
\draw[arc] (agent.15)  to[bend left=20]  (mcp.165);
\draw[arc] (mcp.195)   to[bend left=20]  (agent.345);
\draw[barc] (mcp.east) -- (sage.west);
\end{tikzpicture}
}
\caption{Overview of the system: the ReAct agent (a frozen LLM) receives the \texttt{graph6} string of the target graph, interacts with SageMath through the \texttt{mcp-sage} MCP server, and returns a verified algebraic construction.}%
\label{fig:system}
\end{figure}
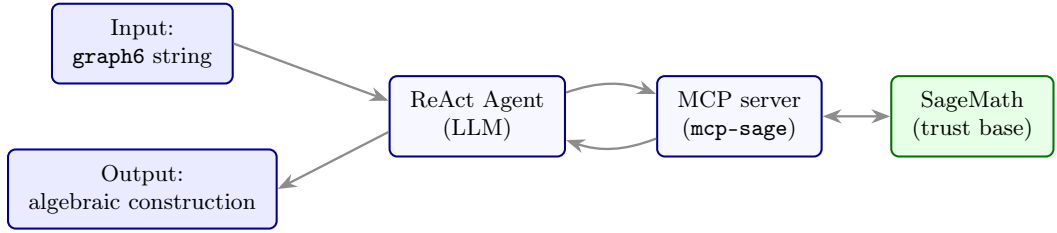

We observe a strong asymmetry.
Finding a candidate construction is costly as it involves searching a large open hypothesis space.
In contrast, verifying a candidate construction is cheap as it requires just one isomorphism test.
We facilitate the construction discovery by means of an agentic loop in which a language model proposes constructions and utilizes the computer algebra system Sage to test properties of the graph and certify candidate constructions (\figureref{fig:system}).

Inside the agentic loop we use a frozen, general-purpose language model without any fine-tuning or per-target adaptation.
It runs as a ReAct agent~\citep{ReAct2023} where it interleaves reasoning with Sage actions.
The agent decides on its workflow and on the sequence of actions, which include analyzing the target (computing the automorphism group, the orbits, and the induced subgraph on each), hypothesizing a construction from that structure, generating Sage code that builds the graph, and checking the result against the input graph.
The agent might refine a hypothesis across several attempts, abandon one for another, backtrack to re-analyze, or probe an auxiliary question such as taking the complement of the graph.

Each target $G$ is presented as a fixed feature record. This record consists of the \texttt{graph6} encoding together with $n$, regularity, $|\Aut(G)|$, and the orbit sizes.
All four are invariants the agent could compute from the encoding, so supplying them saves a call and adds no information; nothing about the target's origin, its membership in a named family, or the construction that produced it is disclosed.
A system prompt for the agent explains the vocabulary of $\Lang$, the acceptance criterion, and the \texttt{graph6} fallback.
Whether a construction lies within the template grammar $\Templ$ is determined independently and not shown to the agent.

The agent interacts with Sage through an MCP server that we contribute as a general-purpose interface. It exposes \texttt{sage\_exec} (run a snippet and return its result, standard output, and any error) plus four session-management operations.
The agent is restricted to these tool calls, with no shell or file-system access, so every computation occurs inside Sage. The MCP server is independent of the graph-construction tasks studied in this paper, and we release it for reuse by any MCP-compatible agentic system.

We verify the construction proposed by the agent by running Sage's \texttt{is\_isomorphic} test against the input graph; the \texttt{Graph(g6($G$))} fallback is recognized by its syntax.

We used a lightweight agent harness, released as \texttt{mcp-minion}, which wraps a language model and has access to any MCP server provided.
For our experiments, we connect it to OpenAI GPT-5.5 (\texttt{gpt-5.5}, released 2026-04-24) via OpenRouter (access date 2026-05) at temperature $T = 0$.
We selected this model from four candidates in a separate pre-study on $40$ graphs (\sectionref{sec:backbone}).
Each run can use at most $80$ agentic steps and $500{,}000$ tokens (input plus output; all token counts in this paper are such totals).
The token cap was introduced after the pre-study of \sectionref{sec:backbone} and was never reached in any run.
\section{Experiments}
\label{sec:experiments}

We run experiments on a $100$-graph benchmark of highly symmetric graphs, a comparison of different LLMs on a $40$-graph pre-study, an evaluation on $50$ random regular graphs, stratified by the number of automorphisms, and a twenty-graph recall test where we selected graphs with planted constructions.
Finally, we run the agent on our new counterexample to the Bernhart--Kainen conjecture.

\subsection{Benchmark population}
\label{sec:bench-pop}

Not every graph has an elegant algebraic construction.
It is usually assumed that whether a graph has such a construction is correlated with how symmetric the graph is, i.e., how many automorphisms it possesses.
Running our agent on a graph that does not have an algebraic construction is an impossible task.
Hence, for the main experiment, we focused on a hundred highly symmetric graphs for which we assume that they possess such constructions.
We utilized the catalog of over seven million two-orbit graphs on up to twenty-five vertices that we published recently~\citep{seka2026enumeratingtwoorbitgraphs}.
From those, we only considered regular non-joins, since graphs that are joins have a trivial construction.
However, being a non-join forces $\gcd(a,b) > 1$ for the orbit sizes $a$ and $b$~\citep{seka2026enumeratingtwoorbitgraphs}.
Hence, we dropped graphs where the number of vertices is a prime, as those are all joins.
We start at $n = 12$: smaller orders have very few regular non-joins (none for $n \le 7$, two each at $n = 8$ and $9$, and eighteen at $n = 10$), and graphs this small are easy to identify directly.
From each admissible order $n \in \{12, 14, 15, 16, 18, 20, 21, 22, 24, 25\}$, we selected the ten graphs with the largest number of automorphisms, giving $10 \times 10$ target graphs in total.
Each order's ten graphs were fixed before the corresponding agent runs; the $n = 24$ order was added by the same rule once its enumeration became feasible.

\subsection{Capability}
\label{sec:capability}

Given only the raw data of the graph, the agent returned a verified non-fallback construction for each of the hundred target graphs.
We analyzed the constructions and identified four recurring families, each with a distinct algebraic signature.
We write $O_1 \oplus O_2$ for a construction in which the first orbit induces $O_1$ and the second induces $O_2$.
In such a construction, the cross-edges are supplied by a stated rule and not edge by edge.

The largest family, with 35 targets spread across all even orders from $n = 12$ to $24$, partitions one orbit into two halves, each joined completely to one part of a bipartite second orbit, as in $C_4 \oplus K_{4,4}$.
A second family, with 17 targets again across the even orders, sets cross-edges by a parity predicate that crosses standard product boundaries, as in $Q_3 \oplus 2K_4$ coupled by Hamming weight mod $2$.
A third family, with 19 targets confined to $n = 15$ and $21$, generalizes crown graphs to three $\mathbb{Z}_3$-indexed parts with cyclic avoidance rules, as in $3K_3 \oplus \overline{K_6}$ sending $T_i$ to $B_{j \neq i}$ and realizing the automorphism group as a wreath product.
The fourth family, with 10 targets all at $n = 25$, uses $D_5$-symmetric constructions built on the $5$-cycle, such as $C_5[K_3] \oplus C_5[\overline{K_2}]$.
The remaining nineteen constructions are inhomogeneous, but still structured.
The group holds nine \emph{apex-over-cliques} constructions (two cliques indexed by $\mathbb{Z}_2$, each carrying apex vertices joined to the whole clique, the apexes typically matched across the two blocks), six weighted blow-ups of a small Cayley base ($\Cay(\mathbb{Z}_6, \{+1, -1\})$ or the triangular prism) with fiber sizes forced by regularity and supported on a distinguished subgroup, three incidence constructions over $\mathbb{F}_2$ or $\mathbb{Z}_4$, and one lexicographic product $C_6[K_2]$ augmented by a vertex per antipodal pair.
To exemplify the agent's range, let us look at one individual construction.
The target graph has $21$ vertices, is $14$-regular, and has orbit sizes $6 + 15$.
The first orbit consists of the six directed edges of the complete graph $K_3$, where two arcs are adjacent unless they are reverses of each other, which gives the octahedron $K_{2,2,2}$.
The second orbit consists of three fibers of five vertices each, indexed by $\mathbb{Z}_3$.
The fibers are pairwise completely joined; there are no edges inside a fiber.
Finally, each directed edge is joined to every fiber except the one at its tail.
As a result, the automorphism group of the graph is the wreath product $S_5 \wr S_3$ of order $(5!)^3 \cdot 3! = 10{,}368{,}000$.
This is visible in the construction since the three fibers permute independently and $\mathbb{Z}_3$ relabels everything.

\subsubsection{Grading construction quality}
\label{sec:human-calib}

To quantify how mathematically interesting and algebraically elegant a construction is, we grade each construction on a zero-to-five canonicity rubric with an LLM judge.
Here five refers to a construction that cleanly combines named families and standard operations.
Zero refers to the fallback with an explicit construction edge by edge.
As a judge, we used Claude Opus 4.7 (released 2026-04-16), run at $T = 0$ with a fixed prompt.
Among the hundred constructions, $47$ received grade $5$, $44$ grade $4$, and $9$ grade $3$, with a median of $4$ and a mean of $4.38$.
The fallback was not used at all.
We compared the LLM judge with a human judge on a $25$-graph subset (hand-labeled by one author).
The subset was drawn from the original $90$-graph benchmark, before the $n = 24$ extension.
The constructions in the subset were presented as LaTeX-rendered notation rather than raw Sage.
The judge is systematically $0.32$ grades stricter.
But $24/25$ of its grades differ from the human grade by at most one, $14/25$ are exact, and the mean absolute error is $0.48$.
The LLM judge can be considered a reasonable proxy for a human judge.
Grades below five identify constructions that are correct but do not achieve full algebraic elegance.
The typical pattern, seen in the audit, is that while some elements are canonical, some parts are fixed by ad-hoc choices.
Such gradations in the quality of a construction cannot be tested by the isomorphism test and tend to rely on some subjective evaluation.

The MCP server\footnote{\url{https://github.com/szeider/mcp-sage}} and the lightweight ReAct agent\footnote{\url{https://pypi.org/project/mcp-minion/}} are released independently as open-source tools.
We also release the system prompt, the grading prompt, the run transcripts, and the per-experiment data.%
\footnote{\url{https://doi.org/10.5281/zenodo.21850460}}

\subsubsection{Comparison with systematic baselines}
\label{sec:baselines}

We compare the agentic constructions with two baseline methods.
The first baseline, \emph{literature lookup}, checks each target graph against a database of $121{,}000$ named graphs (House of Graphs~\citep{BrinkmannEtAl2013,CoolsaetDhondtGoedgebeur2023} and strongly regular catalogs) and runs recognition checks for circulants, strongly regular graphs, Kneser graphs, named graphs such as Petersen and Paley, line graphs, lexicographic products, and complements of circulants.
The second baseline, \emph{strong template enumeration}, enumerates all constructions in a fixed grammar: base components (complete bipartite and multipartite graphs, cycles and their complements, triangular prisms, cliques) combined by a fixed set of construction patterns with cross-edge rules, testing each candidate against every target.

The literature lookup could not identify any of the hundred graphs.
Strong template enumeration found twenty-one of them, all sharing a $K_{k,k}$-core motif (with a small added component such as $C_4$ or a prism, joined by simple split cross-edges).
These twenty-one graphs are also among the graphs with the largest number of automorphisms for their vertex count.
The remaining seventy-nine graphs require constructions outside the predefined grammar.
It is possible that a stronger non-LLM competitor would find more constructions, for instance, by means of inductive logic programming.

\subsection{Additional experiments}
\label{sec:additional}

We have three further experiments: (i) about the LLM chosen for the agent, (ii) whether construction quality changes if the graph is less symmetric, and (iii) whether a fallback is due to the nonexistence of a canonical construction or the agent's failure to recover an existing one.

\subsubsection{Backbone dependence and cost}
\label{sec:backbone}

Before fixing the main LLM for our experiments, we ran four candidate LLMs using the same agentic system, prompt, and verifier.
For this, we used forty random regular graphs with sixteen vertices and orbit counts in $\{2, 6, 10, 14\}$.
These graphs contained symmetric as well as almost structureless ones.

\begin{table}[t]
\floatconts{tab:backbones}%
{\caption{Backbone comparison on the 40-graph pre-study, with the harness, prompt, and verifier held fixed. \emph{Solved} counts verified non-fallback constructions; the mean grade is over all 40 targets, where runs that produced no construction score $0$ and all other runs carry their judge grade.}}%
{\small\begin{tabular}{@{}lrrrr@{}}
\toprule
Model & Solved & Mean grade & Tokens & Cost \\
\midrule
GPT-5.5 & $38/40$ & $\mathbf{3.00}$ & $1.9$M & \$$15.53$ \\
Gemini 3 Flash & $31/40$ & $2.27$ & $9.1$M & \$$5.62$ \\
Gemini 3.1 Pro & $21/40$ & $1.80$ & $8.8$M & \$$22.65$ \\
Claude Sonnet 4.6 & $13/40$ & $1.48$ & $60.8$M & \$$201.95$ \\
\bottomrule
\end{tabular}}
\end{table}

\tableref{tab:backbones} shows the results of the study.
It indicates that performance is not strongly correlated with price, and GPT-5.5 stands out as the clear winner within this group.

The method is inexpensive at this scale.
The average run on the 100-graph benchmark uses $27{,}000$ tokens and $6$ Sage calls, and the maximum is $108{,}000$ tokens and $14$ calls.
The per-run cap of $500{,}000$ tokens was therefore never approached, and no target was decided by exhaustion of budget.
The full run on the hundred graphs cost about \$$23$.

\subsubsection{Symmetry-tracking grade}
\label{sec:reliability}

We run another experiment on fifty random regular graphs with sixteen vertices.
This experiment is intended to see how construction quality depends on the symmetry of the target graph.
Therefore, we categorize these graphs by their number of automorphisms and put them into four bins.
The grades of the LLM judge fall monotonically from $5.00$ over the ten graphs with $|\Aut| \geq 64$ to $3.25$ over the eight in $8$--$63$, $2.50$ over the twenty-two in $2$--$7$, and $2.30$ over the ten asymmetric ones.
All ten graphs in the most symmetric bin are complements of disjoint unions of cycles, so this bin is also structurally homogeneous.
The agent found proper constructions for forty-seven of the fifty graphs and used the fallback construction for three.
The three fallbacks appeared in the two bins of lowest symmetry.

\subsubsection{Recall on planted constructions}
\label{sec:recall}

The final experiment addresses the question whether the agent misses constructions for graphs for which constructions exist.
In that case, a fallback can be considered a failure of the agent.
Therefore, we ran the agent on twenty graphs built from planted constructions.
Nineteen of these graphs have compositional algebraic constructions.
One is asymmetric, i.e., has only the trivial automorphism, but is still algebraically defined: it has the vertex set $\mathbb{Z}_{31}$, with $x$ joined to $x + 1$, $x - 1$, and $x^2 + 3x + 7$.
Given only the \texttt{graph6} string, without the symmetry fields (\sectionref{sec:method}), the agent recovered verified constructions for the nineteen compositional graphs; it used the fallback for the asymmetric graph.
It identified some graphs as circulants or Cayley graphs instead of the planted product construction.

\subsection{Bernhart--Kainen case study}
\label{sec:bk-case}
\av{\enlargethispage*{5mm}}

In this section, we report on an application of our agent to construct a graph we found by enumeration.
Here we consider the Bernhart--Kainen conjecture on dispersable book embeddings~\citep{BernhartKainen1979}.
It states that every $k$-regular bipartite graph admits a vertex ordering and a proper edge coloring with $k$ colors such that edges of the same color do not interleave.
The smallest previously known counterexample is the $20$-vertex Folkman graph~\citep{AlamBekosDujmovicGronemannKaufmannPupyrev2021}.
We checked all two-orbit graphs in our recently published catalog~\citep{seka2026enumeratingtwoorbitgraphs} and identified a $5$-regular two-orbit graph on sixteen vertices that is also a counterexample to the conjecture.
It is now the smallest counterexample known.

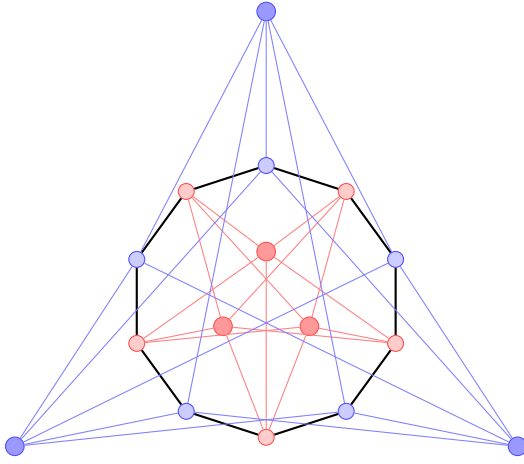
\begin{figure}[t]
\centering
\begin{minipage}[c]{0.4\textwidth}
\centering
\begin{tikzpicture}[scale=1.2,
    cycle_even/.style={circle, draw=blue!70, fill=blue!20, minimum size=6pt, inner sep=0pt},
    cycle_odd/.style={circle, draw=red!70, fill=red!20, minimum size=6pt, inner sep=0pt},
    group_a/.style={circle, draw=red!70, fill=red!40, minimum size=7pt, inner sep=0pt},
    group_b/.style={circle, draw=blue!70, fill=blue!40, minimum size=7pt, inner sep=0pt},
    cycle_edge/.style={thick, black},
    cross_edge_a/.style={thin, red!50},
    cross_edge_b/.style={thin, blue!50}
]

\def\radius{1.5}
\foreach \i in {0,...,9} {
    \pgfmathsetmacro{\angle}{90 - \i*36}
    \pgfmathparse{mod(\i,2)==0 ? "cycle_even" : "cycle_odd"}
    \edef\style{\pgfmathresult}
    \node[\style] (c\i) at (\angle:\radius) {};
}

\foreach \i in {0,...,9} {
    \pgfmathtruncatemacro{\j}{mod(\i+1,10)}
    \draw[cycle_edge] (c\i) -- (c\j);
}

\node[group_a] (a0) at (90:0.55) {};
\node[group_a] (a1) at (210:0.55) {};
\node[group_a] (a2) at (330:0.55) {};

\node[group_b] (b0) at (90:3.2) {};
\node[group_b] (b1) at (210:3.2) {};
\node[group_b] (b2) at (330:3.2) {};

\foreach \a in {a0, a1, a2} {
    \foreach \c in {1,3,5,7,9} {
        \draw[cross_edge_a] (\a) -- (c\c);
    }
}

\foreach \b in {b0, b1, b2} {
    \foreach \c in {0,2,4,6,8} {
        \draw[cross_edge_b] (\b) -- (c\c);
    }
}

\end{tikzpicture}
 \end{minipage}\hfill
\begin{minipage}[c]{0.5\textwidth}
\caption{Smallest known counterexample to the Bernhart--Kainen conjecture: a 5-regular two-orbit graph on 16 vertices. The 10-cycle forms the outer ring, light red and blue marking odd and even positions. Groups~$A$ (dark red, inner) and~$B$ (dark blue, outer) are independent triples; each vertex of~$A$ joins every odd cycle vertex, each vertex of~$B$ every even one.}%
\label{fig:bernhart-kainen}
\end{minipage}
\end{figure}

We provided our agent the \texttt{graph6} string and metadata, and it came up with a verified construction.
It used six tool calls and $27{,}700$ tokens, with a cost of about \$$0.22$.
The found construction builds the graph as $C_{10} \oplus 2\overline{K_3}$: the $10$-cycle $\Cay(\mathbb{Z}_{10}, \{+1, -1\})$ as one orbit, two independent triples ($2\overline{K_3}$) as the other, joined by a parity cross-edge rule that sends each triple to all five cycle vertices of one parity class under the quotient $\mathbb{Z}_{10} \to \mathbb{Z}_2$ (\figureref{fig:bernhart-kainen}): the first triple to the odd cycle vertices, the second to the even. Then each vertex has degree $5$, and the graph is bipartite.
This construction exposes the parity coupling between the cycle and the two triples.
Dispersability can therefore be checked by inspecting the construction rather than by exhaustive search over vertex orderings and page assignments.

\section{Related Work}
\label{sec:related}

\emph{Symbolic graph enumeration.}
There are several methods to compute exhaustive lists of graphs with a given property, recorded as \texttt{graph6} strings.
These methods include canonical augmentation~\citep{McKay1998}, SAT modulo symmetries~\citep{SMS2021}, and SAT+CAS~\citep{BrightEtAl2019}.
Such lists certify existence, but provide no explanation or algebraic interpretation.
This is the input setting that we address, not methods we compete with.
Our benchmark graphs come from such an enumeration~\citep{seka2026enumeratingtwoorbitgraphs}.

\emph{Template- and library-based algebraic discovery.}
The survey by \citet{PerezRoses2014} lists various methods for constructing large degree-diameter graphs.
The most successful ones operate within fixed catalogs of algebraic families, where voltage assignments account for about $60\%$ of the record entries.
The remaining graphs are Cayley and circulant constructions found with computer search that optimizes parameters within a human-supplied family.
\citet{Jooken2025}, by contrast, surveys computer-assisted graph theory more broadly and observes that LLMs have not yet produced breakthrough results in the field.
Our template-enumeration baseline (\sectionref{sec:baselines}) follows this line of research but only recovers a fraction of the constructions found by our agent.

\emph{LLMs and agents for mathematical discovery.}
Language-model-based agents have been used extensively over the last few years for mathematical discovery.
However, they are mainly used for proof construction or object search.
Some prominent examples include the following.
FunSearch~\citep{FunSearch2023} pairs an LLM with an evolutionary program database and a domain-specific scoring evaluator.
AlphaGeometry~\citep{AlphaGeometry2024} uses a transformer model trained on $100$ million synthetic theorems coupled with a forward-chaining deduction engine; it solved math-olympiad geometry problems.
AlphaProof~\citep{AlphaProof2025} emits Lean tactics in a reinforcement-learning loop, again trained on millions of auto-formalized problems, with days of per-target test-time adaptation.
It also solves olympiad-level mathematics problems.
All three place a sound verifier or evaluator in their trust base, as we do.
However, each uses heavy training or per-target adaptation and a domain-specific verifier.
In contrast, our proposer is an off-the-shelf frozen language model, and our verifier decides an exact target.
Also, our approach is not focused on the discovery of new theorems, but on the explanation of objects found by search or enumeration.

\section{Conclusion}
\label{sec:conclusion}

We have established and tested a neurosymbolic pairing of a frozen, general-purpose LLM agent that proposes constructions over an open language and uses an exact symbolic verifier as a trust base.
The agent proposes; the symbolic side verifies.
The agent interacts with a computer algebra system via our MCP interface.
We tested this approach on different graphs and graded the quality (canonicity) of the proposed constructions.

For the main experiment on the hundred symmetric benchmark graphs, the agent could find genuine algebraic constructions for all of them.
Our additional experiments establish the robustness of this approach.
We applied our approach to the new Bernhart--Kainen counterexample, which is now the smallest known, and found an algebraic construction.

Our approach applies to any discrete object beyond graphs.
Our MCP server, which connects agents to the Sage computer algebra system, can be used beyond our focus area of finding and verifying algebraic constructions.
For future research, we suggest a systematic construction of all graphs with ten vertices, starting with a template that gets extended incrementally by new constructions found by the agent.

\acks{This research was funded by the Austrian Science Fund (FWF), grants 10.55776/COE12 and 10.55776/P36688.}

\bibliography{references}

\end{document}